\documentclass[a4paper]{article}

\usepackage[margin=1in]{geometry}
\usepackage[utf8]{inputenc} % allow utf-8 input
\usepackage[T1]{fontenc}    % use 8-bit T1 fonts
\usepackage{url}            % simple URL typesetting
\usepackage{booktabs}       % professional-quality tables
\usepackage{amsfonts}       % blackboard math symbols
\usepackage{nicefrac}       % compact symbols for 1/2, etc.
\usepackage{microtype}      % microtypography

\usepackage{amsmath}
\usepackage{bbm}

\usepackage[normalem]{ulem}

\usepackage{mathtools}

\usepackage{amssymb,amsmath,amsthm}
\usepackage{soul}
\usepackage{url}            % simple URL typesetting
\usepackage{xcolor}
\usepackage{xr-hyper,refcount}
\usepackage[colorlinks]{hyperref}  

\usepackage[utf8]{inputenc} % allow utf-8 input
\usepackage[T1]{fontenc}    % use 8-bit T1 fonts
\usepackage{hyperref}       % hyperlinks
\usepackage{url}            % simple URL typesetting
\usepackage{booktabs}       % professional-quality tables
\usepackage{amsfonts}       % blackboard math symbols
\usepackage{nicefrac}       % compact symbols for 1/2, etc.
\usepackage{amsmath}
\usepackage{graphicx}
\usepackage{algorithm}
\usepackage{algorithmic}
\usepackage{caption}
\usepackage{subcaption}
\usepackage{authblk}
\usepackage{textcomp}

\newtheorem{theorem}{Theorem}
\newtheorem{lemma}{Lemma}

\title{What is Fair? Exploring Pareto-Efficiency for Fairness Constrained Classifiers}
\begin{document}

\author[1,2]{Ananth Balashankar}
\author[2]{Alyssa Lees}
\author[2]{Chris Welty}
\author[1]{Lakshminarayanan Subramanian}
\affil[1]{Courant Institute of Mathematical Sciences, New York University}
\affil[2]{Google AI, New York}
\affil[ ]{ananth@nyu.edu, alyssalees@google.com, welty@google.com, lakshmi@nyu.edu}

\maketitle
\begin{abstract}
The potential for learned models to amplify existing societal biases has been broadly recognized. Fairness-aware classifier constraints, which apply equality metrics of performance across subgroups defined on sensitive attributes such as race and gender, seek to rectify inequity but can yield non-uniform degradation in performance for skewed datasets. In certain domains, imbalanced degradation of performance can yield another form of unintentional bias. In the spirit of constructing fairness-aware algorithms as societal imperative, we explore an alternative: Pareto-Efficient Fairness (PEF). Theoretically, we prove that PEF identifies the operating point on the Pareto curve of subgroup performances closest to the fairness hyperplane, maximizing multiple subgroup accuracy. Empirically we demonstrate that PEF outperforms by achieving Pareto levels in accuracy for all subgroups compared to strict fairness constraints in several UCI datasets.

%As awareness increases of the potential for learned models to amplify existing societal biases, the field of ML fairness has developed mitigation techniques. A prevalent methodology applies constraints, including equality of performance, with respect to subgroups defined over the intersection of sensitive attributes such as race and gender. Enforcing such constraints when the subgroup populations are considerably skewed with respect to a target can lead to unintentional degradation in performance, without benefit to any individual subgroup. 
%In order to avoid such performance degradation, we propose Pareto-Efficient Fairness (PEF), which identifies the operating point on the Pareto curve of subgroup performances closest to the fairness hyperplane. Specifically, PEF finds a Pareto Optimal point which maximizes multiple subgroup accuracies. The algorithm scalarizes using the adaptive weighted metric norm by iteratively searching the Pareto region of all models enforcing the fairness constraint. PEF is backed by strong theoretical results on discoverability and provides domain practitioners finer control in navigating both convex and non-convex accuracy-fairness trade-offs. Empirically, we show that PEF increases performance of all subgroups in skewed synthetic data and UCI datasets.
\end{abstract}

\section{Introduction}

As repeatedly demonstrated in the news, medicine, law and numerous related ML papers  \cite{Zhao2017MenAL, DBLP:journals/corr/abs-1806-04959}  \cite{BolukbasiCZSK16a}, societal inequities have the very real risk of being vastly exacerbated if machine learning algorithms do not explicitly address fairness in model formulation and data collection. Numerous philosophical notions of fairness exist (distributive, procedural, etc) \cite{abstractions, GrgicHlaca2016TheCF} \cite{pac-fair} and the appropriateness of each definition may depend on context. Theoretically, in an equitable world of perfect data, implying perfect accuracy across all possible subgroup populations, a uniformly fair classifier may be created. However, with skewed real-world data, \cite{Menon2018TheCO} has shown that a tradeoff exists between fairness and accuracy. We propose an alternative fairness constraint based on Pareto-Efficiency \cite{Godfrey2007} to avoid performance degradation within some subgroups, while striving for increased accuracy for \textit{all} subgroups.

A popular approach to learn fairness-aware classification models is to enforce strict metric equality constraints in relation to sensitive variables such as race and gender. Such equality constraints are being adopted as law in some locals \cite{NYCLaw}. However, by definition, enforcing strict equality constraints will ensure classification accuracy is limited by the worst performing subgroup. Due to a variety of reasons, historical injustices \cite{criminal}, sampling bias \cite{DBLP:journals/corr/ChakrabortyMBGG17}, selection bias \cite{pmlr-v81-speicher18a} among others, subgroup populations are often not fully or fairly represented in commonly used real-world datasets. The discrepancies are particularly alarming when algorithmic models are used to predict medical and legal outcomes. For example, in a cardiology study of over 4000 ER patients with cardiac event symptoms \cite{Allabban},  \textit{no symptoms} were found to be predictive of a heart attack in white women. In black males, only an unrelated symptom (diaphoresis) was found to be indicative of a future cardiac event with 95 percent confidence, while in white males relevant features (left arm radiation, pressure/tightness) were detected with high accuracy. To reiterate, a classifier built on this longitudinal and ``diverse'' dataset to predict an ER cardiac event, could conceivably achieve high accuracy for white men, while only yield trivial accuracy of 50 percent for white women and black men.

\begin{figure}[t]
\centering
		\includegraphics[width=2.5in,height=1.7in]{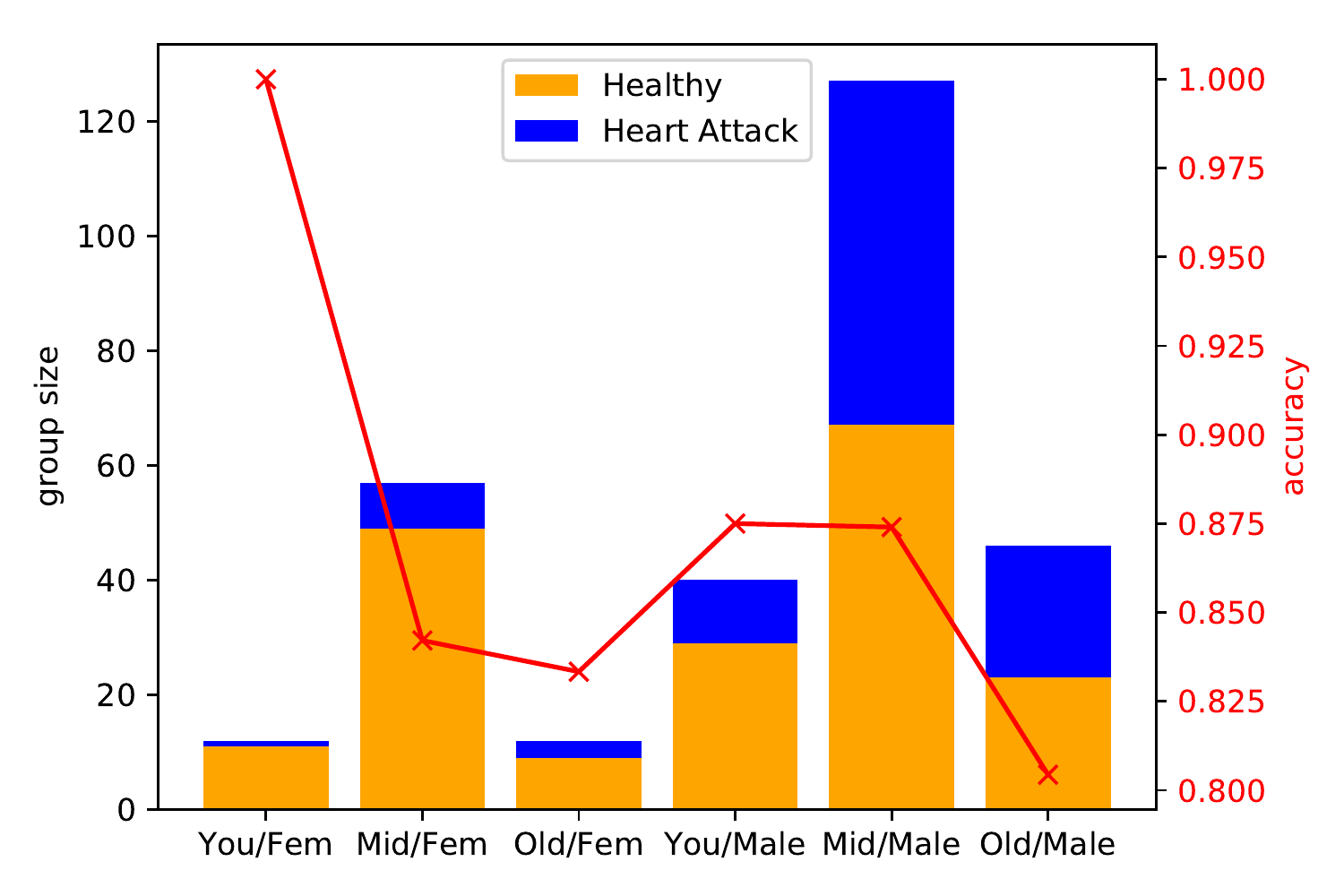} 
		\setlength{\belowcaptionskip}{-8pt} 
		\caption{Illustration of group level skew on UCI Heart Attack dataset: Groups based on age and gender have disparate target distribution and applying strict equality might unintentionally degrade accuracy without benefitting any group.}	
		\label{heart-attack-disparity}
\end{figure}
\begin{figure}[h]
\centering
		\includegraphics[width=2.5in,height=1.7in]{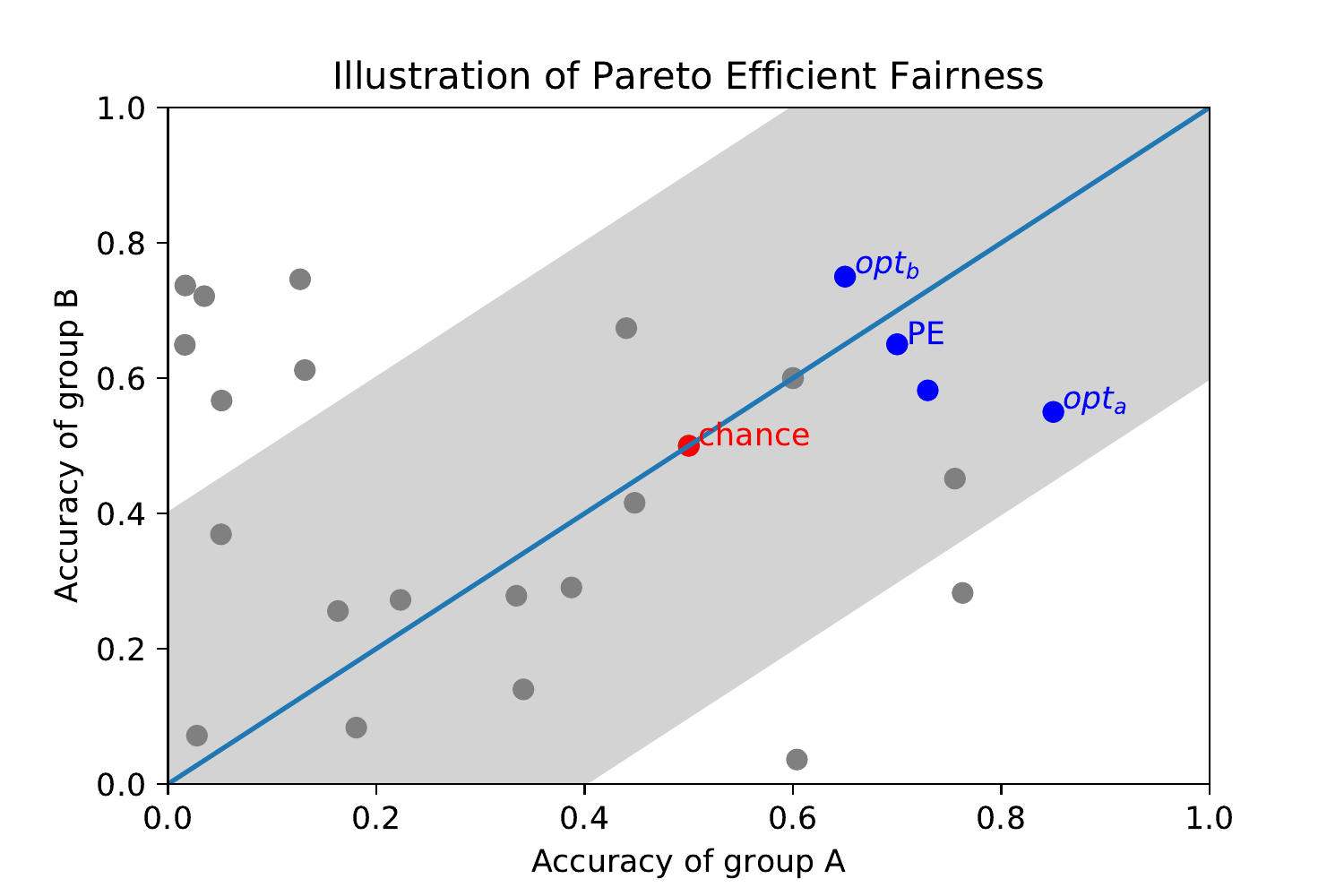} 
		\setlength{\belowcaptionskip}{-8pt} 
		\caption{Illustration of Pareto Efficiency Fairness on synthetic data: The most accurate and strictly-equal classifier lies on the operating point of (0.6, 0.6). If accuracy for each of the groups is separately maximized, we would choose points $opt_a = (0.83, 0.55), $ and $opt_b = (0.63, 0.77)$ respectively. However, PEF will choose among the points on the Pareto-front, the point $PE = (0.71,0.63)$, which improves accuracies of both $a$ and $b$ within the fairness relaxation bounds (in gray).}
		\label{pareto-explain}
\end{figure}

A philosophical question for the reader and practitioners is - should a classifier be constructed in such a scenario? Without full historical data and appropriate domain causal knowledge, it may be infeasible to approach *fair* learning \cite{Madras:2019:FTC:3287560.3287564}.  The above example is extreme both in repercussions and the skew of the data. However, we argue and demonstrate that such skew is common in frequently used datasets to evaluate fairness aware learning - UCI Adult, German credit and Heart attack. Figure \ref{heart-attack-disparity} demonstrates the skew in accuracy by age-gender subgroups in the UCI Heart Attack dataset. 

In scenarios where the domain practitioners deem the above considerations acceptable for constructing a classifier, we propose an alternative fairness constraint based on Pareto-Efficiency \cite{Godfrey2007} to avoid the unintentional degradation in subgroup performance, while striving for increased accuracy.  The methodology is applicable to highly skewed data as described in both cardiac examples above. The Pareto-Efficient Fairness (PEF) constraint restricts the choice of ML models to the Pareto frontier to ensure higher accuracy across *fair* model options. In some cases, a Pareto-Efficient definition may be at odds with a strict equality fairness criterion. Figure \ref{pareto-explain} illustrates cases on a synthetic dataset where extremely unequal models might be Pareto optimal and vice versa. However, PEF avoids this pitfall by limiting the search space on the Pareto frontier within acceptable fairness bounds. 

Our proposed bias loss function achieves Pareto-Efficient performance, which outperforms solutions based on equalizing subgroup performance. Our algorithm iteratively searches for Pareto-optimal subgroup performance by leveraging the benefits of transfer learning to minimize the Group Pareto loss. Using theory from multiple objective optimization for continuous Pareto fronts, we prove that if the data distribution has high disalignment \cite{Menon2018TheCO} between the subgroup and the outcome, PEF will discover all Pareto optimal points and converge to a solution that is better than the Bayes optimal solutions for existing fairness constraint based algorithms (if it exists). Empirically, we show that our approach achieves an operating point which is better both in terms of global accuracy and individual subgroups accuracy than methods which approximate hard constraints of equality \cite{Zhao2017MenAL} and adversarial multi-task learning \cite{Beutel2017DataDA} on three UCI datasets.

\section{Pareto-Efficient Fairness}

\subsection{Motivation}
To motivate the need for Pareto-Efficient Fairness, consider a very simplistic binary classification task $X \rightarrow \{0,1\}$, where $X \in \mathbb{R}$ is a continuous scalar feature for each example in the dataset D. We partition the dataset D into a set of groups $G = \{a, b\}$ such that the examples in groups $a$ and $b$ differ in their values for sensitive variables S.

We denote the accuracy of a classifier $h \in H$ evaluated on test samples from groups a and b as $f_a$ and $f_b$ respectively. We thus say that the classifier $h$ evaluates to an operating point $(f_a, f_b)$. We can evaluate various operating points for classifiers $h \in H$ by varying values of the threshold $t$ to define the scatter plot of operating points, if they take the form
\begin{equation}
h(X)=
\begin{cases}
0, & \text{if}\ X<t \\
1, & \text{otherwise}
\end{cases}
\end{equation}
% . 

% \begin{figure}[h!]
% 	\begin{center}
% 		\includegraphics[width=3.0in,height=1.7in]{pareto-explain.pdf} 
%         \setlength{\belowcaptionskip}{-8pt} 
% 		\caption{Illustration of Pareto Efficiency Fairness: The most accurate and strictly-equal classifier lies on the operating point of (0.6, 0.6). If accuracy for each of the groups is separately maximized, we would choose points $opt_a = (0.83, 0.55), $ and $opt_b = (0.63, 0.77)$ respectively. However, PEF will choose among the points on the Pareto-front, the point $PE = (0.71,0.63)$, which improves accuracies of both $a$ and $b$ within the fairness relaxation bounds (in gray).}
% 		\label{pareto-explain}
% 	\end{center}
% \end{figure}

In Figure~\ref{pareto-explain}, we plot these operating points $(F_a, F_b)$ over a given data distribution by varying $t$. If we assumed that the data is drawn from a uniform random distribution over a class-balanced label set for both groups $a$ and $b$ and the label is generated by flipping a fair coin, then the expected operating point for any classifier would be (0.5, 0.5) (denoted by ``chance''). In other non-degenerate scenarios, it is possible to maximize the accuracy ($> 0.5$) of a single group, regardless of the accuracies of the other groups. In our case, we denote such operating points that maximize accuracy of groups $a$ and $b$, by $opt_a$ and $opt_b$ respectively. A strict equality fairness criterion would require that the classifier operate on the $x=y$ line.

However, if the objective is to improve the performance of all groups to meet the levels of the highest performing groups, then choosing points on the line $x=y$ might not be desirable. Such objectives are common in policies around affirmative action \cite{doi:10.1177/1043463192004002004} and recent works in fairness literature \cite{pmlr-v81-buolamwini18a}, where even though parity between demographic groups is desired, interventions are made to improve all groups to the level of the highest performing group. Hence, choosing the Pareto-Efficient point (denoted by ``PE'') may be a more desirable solution as it will increase the accuracy  of \textbf{both} groups $a$ and $b$ compared to a solution obtained by enforcing a strict equality constraint. 

\subsection{Definitions}
For completeness, we provide some definitions of model performance metrics which may be used to evaluate fairness on the held-out test data set.

\textbf{Accuracy}: The fraction of test samples which were classified correctly by the model as compared to the ground truth class labels.

\textbf{False Positive Rate}: In case of a binary classification task with positive and negative classes, this is the fraction of test samples which were incorrectly assigned positive by the model among the total number of ground truth negative samples.

\textbf{False Negative Rate}: In case of a binary classification task with positive and negative classes, this is the fraction of test samples which were incorrectly assigned negative by the model among the total number of ground truth positive samples.

These simple definitions of model performance have been used as an objective for maximization/minimization under constraints such that sensitive group level model performance (could be measured using a different metric than the one used as an objective) be equal.

\textbf{Parity Loss}: If the group level model performance are unequal, then the sum of absolute deviation of the group level performances from the overall model performance is defined as the Parity loss. If $f_g$ denotes the model performance of group $g$ and $f$ denotes the overall model performance, then parity loss is given by
\begin{align}
    \Sigma_g | f_g - f|
\end{align}

There are many fairness definitions proposed \cite{DBLP:journals/corr/abs-1808-00023} in literature, but we present one of the commonly used strict-equality constraints below followed by our definition of Pareto-Efficient Fairness.

\textbf{Equality of Odds. \cite{HardtPS16}:}
We say that a predictor $\hat{Y}$ satisfies equalized odds with respect to the sensitive attribute set $A$ and outcome $Y$, if $\hat{Y}$ and $A$ are independent conditional on $Y$.
\begin{align}
    P(\hat{Y}=\hat{y}|Y=y, A=m) = P(\hat{Y}=\hat{y}|Y=y, A=n), \forall y, \forall m,n \in A
\end{align}

\textbf{Pareto-Efficiency}: We define Pareto-Efficient points as the set of operating points for which there does not exist another point, which has better performance (e.g. accuracy) across all the groups. 
\par
One possible pitfall of the above definition for fairness is that points $opt_a$ or $opt_b$ could be selected as a Pareto-Efficient solution. Both are trivially Pareto-Efficient points, since there are no other points which performs better across all groups, but are trivially unequal across groups.

\textbf{Pareto-Efficient Fairness}: We say an operating point is Pareto-Efficient Fair if it is Pareto-Efficient and minimizes the variance of the Pareto error across groups.
\par 
Formally, for a set of thresholds characterizing Pareto-Efficient points: $T_{PE}$, performance metric for $g$: $f_g$, optimum performance metric across all operating points for group $g$: $f_{opt-g}$, the Pareto error for group $g$: $\epsilon_g \in \mathcal{E}_G$, variance across groups all $g \in G$: $\sigma_G^2$, we intend to find a threshold $t_{PE-fair}$ that characterizes a Pareto-Efficient Fair operating point.
\begin{align}
\epsilon_g &= 1 - \frac{f_g}{f_{opt-g}}\\
t_{PE-fair} &= \underset{T_{PE}}{\text{arg min }} \sigma_G^2 (\mathcal{E}_G)
\end{align}

Since it is empirically difficult to find all Pareto-Efficient thresholds $T_{PE}$ without sufficient exploration, a simple heuristic is to choose a threshold that minimizes the total absolute Pareto-penalty.
\begin{align}
t_{PE} &= \underset{T}{\text{arg min }}  \Vert \mathcal{E}_G \Vert_1
\end{align}
We combine these two minimization criterion using a Lagrangian factor $\alpha$ as follows:
\begin{align}
t_{PE-fair} &= \underset{T}{\text{arg min }} \alpha \Vert \mathcal{E}_G \Vert_1 + (1 - \alpha) \sigma_G^2 (\mathcal{E}_G)
\end{align}
% The choice of $\alpha$ is domain dependent and can be set to trade off the deviation from the group's heuristic pseudo-optimum reference performance ($\alpha$) and the variance in the deviation across groups ($1 - \alpha$). 
This formulation should not be confused as $\alpha$ to be simple fairness and accuracy tuning parameters, as proposed in \cite{Zhao2017MenAL}. The nuance is in the fact that, $\alpha$ is used relative to the heuristic pseudo-optimum performance of groups, which is central to the argument of Pareto-Efficient Fairness, thereby encouraging all subgroups to perform at their best possible levels. This is demonstrated when $\alpha = 0$ and $\sigma_G^2(\mathcal{E}_G)$ is minimized, which is not the same as equality of odds \cite{HardtPS16}. Similarly, when $\alpha = 1$, we minimize $\Vert \mathcal{E}_G \Vert_1$, which is a different than the unconstrained optimization in \cite{Zhao2017MenAL}

\textbf{Group Pareto Loss:} 
We now generalize the simple example above to any binary classification task. Here, the minimization criterion of the Group Pareto Loss in equation (5) holds, but instead of determining a specific threshold corresponding to a Pareto-Efficient operating point, we minimize the group Pareto Loss over the parameters of the binary classification model.
% for dataset $D$ with $n$ variables.
% $m$ of which are considered sensitive or protected with one label in a classification task. 
% In this scenario, arriving at a point of equity across all groups $(O(2^m))$ defined over m sensitive variables is difficult. Trying to achieve this strictly may lead to the collapse of the model to random chance (``chance'' point in Figure~\ref{pareto-explain}) where the constraint of equity is satisfied trivially. Instead of choosing an approximate operating point that satisfies a constraint of an equal performance metric in this high dimensional space $O(2^m)$, we  aim to choose a Pareto-Efficient operating point. The Pareto-Efficient operating point, by definition will perform \textit{better}, if not equally across all groups' performance metrics as compared to an operating point chosen by any approximation algorithm trying to achieve strict equality.

% the objective of equalized odds is to learn a model which performs equally well (be it any of accuracy, AUROC, false positive rate, true positive rate, etc) on each of the groups. With a complementary objective of achieving the highest overall performance, we would expect to achieve a trade-off between the two with some groups benefiting in performance at the cost of degradation for the others. 

\begin{figure}[h]
\centering
		\includegraphics[width=2.5in,height=1.7in]{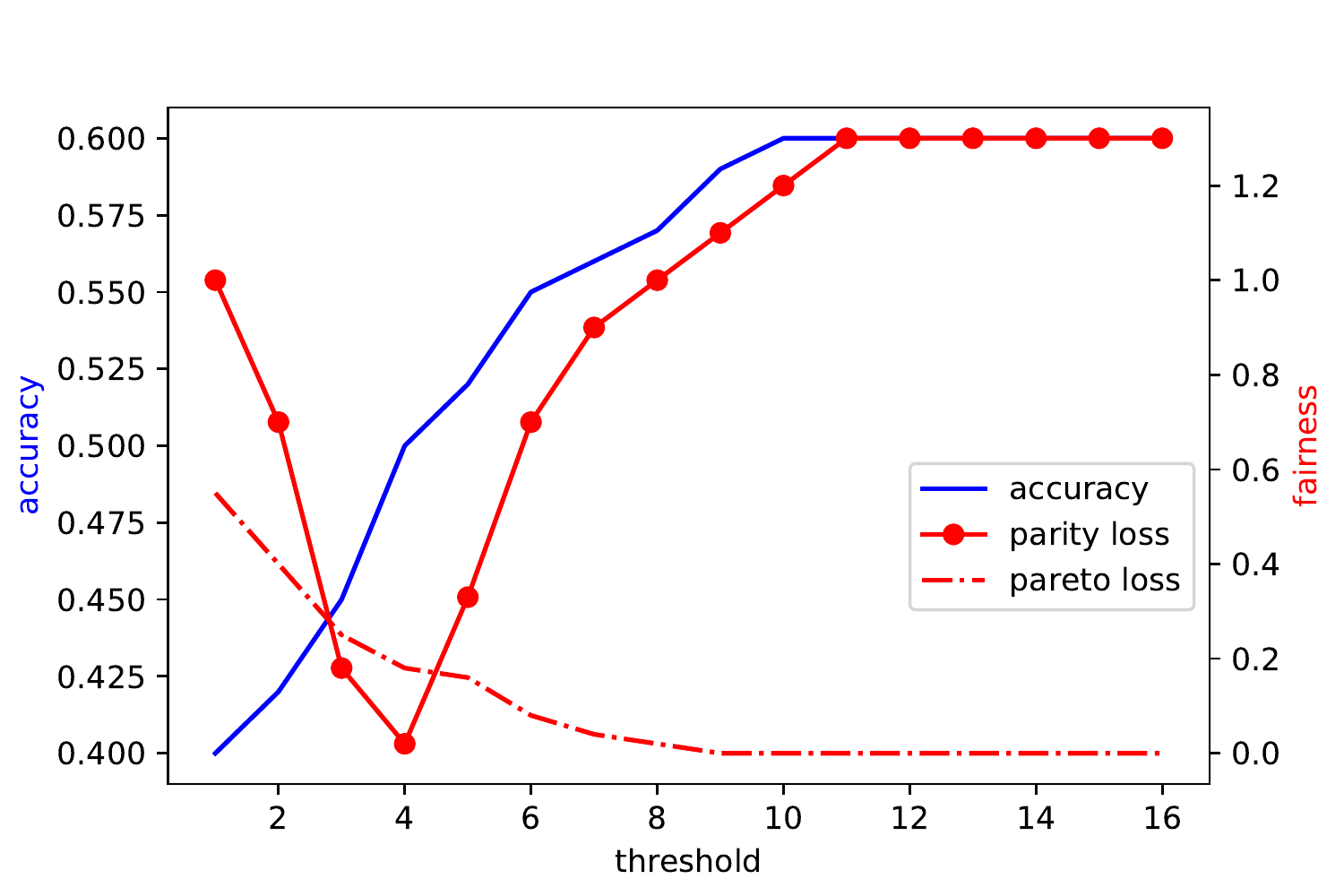} 
        \setlength{\belowcaptionskip}{-8pt} 
		\caption{Illustration of preference of Pareto loss over Parity loss. In this synthetic data scenario, two groups perform at random accuracy level, regardless of the threshold chosen and two other groups have higher accuracies when the thresholds are increased. The Pareto loss depicts how far each of the groups are from their corresponding optimal accuracy levels. Parity loss depicts the discrepancy between the groups accuracies. Parity loss is minimized when all groups perform at random accuracy levels, whereas Pareto loss is minimized when all groups achieve their optimal accuracies and hence is a better alternative.}
		\label{pareto-parity}
\end{figure}
\begin{figure}[h]
\centering
		\includegraphics[width=2.5in,height=1.7in]{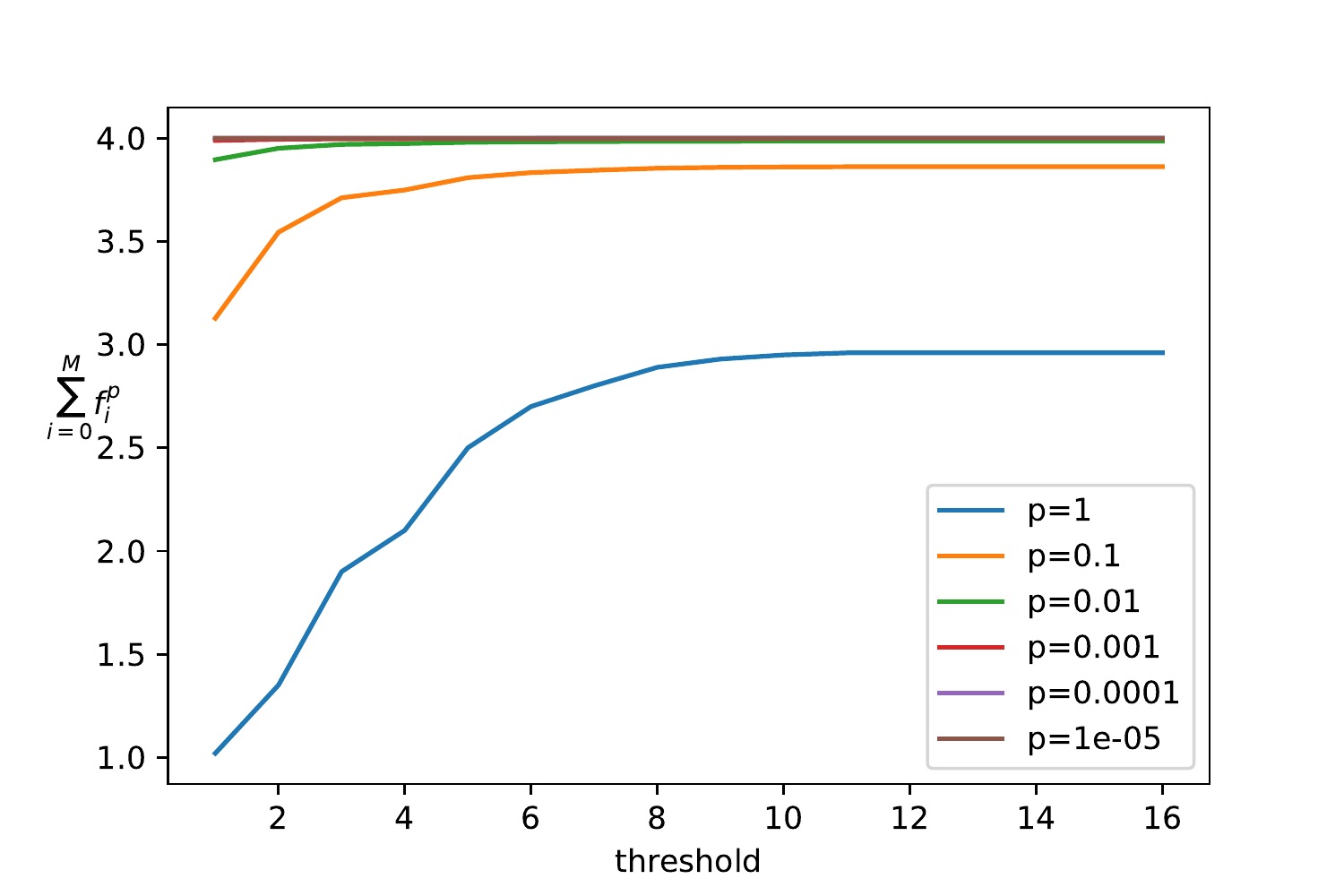} 
        \setlength{\belowcaptionskip}{-8pt} 
		\caption{Pareto geometry condition is satisfied for discoverability of all Pareto-Optimal points for Pareto-Efficient Fairness for low values of p as required by Lemma 3.3}
		\label{pareto-geometry}
\end{figure}

%Since each group tries to achieve its own "optimal" performance during the course of the bias mitigation, this can be thought of achieving Pareto-Efficiency, where we try to optimize for overall performance, while accepting decrease in performance for certain groups if and only if there is at least one group which benefits as a result.

% In order to achieve desirable results, we want to ensure that any penalty in performance is not dominated by only certain groups, but shared across groups equitably. These two constraints can be thought of as constraining the initial performance optimization objective while
% minimizing the full Pareto loss $L_{p}(o, \hat{o})$. The bias mitigation loss can be specified as

% \[ L_{b}(o, \hat{o}) =\min \sum_{g \in G}{L(g)} \]
% where 
% \[ L(g) = max(0, F_{opt}[g] - F[g]) \]

% subject to
% $L(g) < \epsilon$ for all groups $g \in G$ for some $\epsilon > 0$,
% where G is the set containing all possible groups over the set of sensitive variables $S$, $F_{opt}[g]$ is the heuristic pseudo-optimal evaluation score for group $g$ among currently observed operating points and $F[g]$ is the actual evaluation score for a given model.

The Group Pareto Loss is augmented with an appropriate loss weight ($\lambda$) through the Lagrangian dual formulation similar to \cite{Eban2016}, along with the standard cross entropy classification loss: $\mathcal{L}_{ce}$ \cite{cross-entropy}, to yield the Pareto-Efficient Fairness Loss: $\mathcal{L}_{p}$. Note that the standard cross entropy classification loss aims to maximize overall performance, whereas the penalty term weighted by $\lambda$ is used to ensure that such a maximum overall performance should be achieved while minimizing Pareto loss. In scenarios, where the maximum overall performance is achieved when each group's performance is also maximized, then we would not need this augmentation. For cases, where such Pareto optimal and overall optimal operating points do not coincide, we are able to denote our preference between these two operating points using the penalty term $\lambda$.

\begin{align}
    \mathcal{L}_{p} = \mathcal{L}_{ce} + \lambda(  \alpha \Vert \mathcal{E}_G \Vert_1 + (1 - \alpha) \sigma_G^2 (\mathcal{E}_G))
\end{align}

\subsection{Pareto-Efficient Algorithm}
We now put the components mentioned above together and present the Pareto-Efficient bias mitigation algorithm (Algorithm \ref{alg:algorithm-pareto}). This algorithm is an in-processing algorithm which trains a joint model to have favorable fairness properties, as opposed to a post-processing algorithm which uses a pre-trained model for fine-tuning \cite{pmlr-v65-woodworth17a}. In order to obtain a heuristic pseudo-optimal group accuracy $f_{opt-g}$ for each group $g$, we train the classifier $M_g$ to minimize $\mathcal{L}_{ce}$ on samples in group $g$ from dataset $D$. Although this may not be the best estimate of a group population's true optimal, we avoid transfer learning during bootstrapping as it benefits larger groups more than smaller groups \cite{PapernotAEGT16}. Instead, we propose an iterative approach where $f_{opt-g}$ is updated in each iteration if a better group accuracy is achieved by a jointly trained model. 
% In the evaluation, we present results from the heuristic initialized with the separately trained groups. 
We use these heuristic pseudo-optimal accuracies to jointly train a Pareto-Efficient model $M$ on all subgroups to minimize the $\mathcal{L}_p$ in every batch. We further strictly ensure that the mini-batch is representative of the group distributions by sampling group-wise batch samples proportionately.

\begin{algorithm}
\caption{Iterative Pareto-Efficient Bias Mitigation}
\label{alg:algorithm-pareto}
\begin{algorithmic}
\STATE{$G$: set of sensitive groups, $D$: dataset, $D_g$: data of group $g \in G$}
\FOR{$g \in G$}
\STATE{$M_g = \text{arg min }\mathcal{L}_{ce}(D_g) $}
\STATE{$f_{opt-g} = \text{eval}(M_{g}, D_{g})$}
\STATE{$f_g = \emptyset$}
\ENDFOR
\WHILE{$\exists g \in G, f_g = \emptyset \lor f_g > f_{opt-g}$}
\STATE{$f_{opt-g} = \text{max}(f_g, f_{opt-g}), \forall g \in G$}
\STATE{$M$ = \text{arg min }$\mathcal{L}_p(D)$}
\STATE{$f_g =$  \text{eval}($M$, $D_{g}$), $\forall g \in G$}
\ENDWHILE
\end{algorithmic}
\end{algorithm}

Our proposal aims to achieve ``potentially optimal'' performance for each of the groups while performing better than approximate fairness constrained classifiers.
% and that at a minimum a classifier aiming towards \textit{fairness} should be able to reflect the underlying sampled sub-populations distributions as accurately as possible. 
In Figure~\ref{pareto-parity}, we illustrate
% why minimizing the Group Pareto loss achieves this instead of the Parity loss enforced by strict equality constraints \cite{Zhao2017MenAL}, 
this by applying Algorithm \ref{alg:algorithm-pareto} on a synthetic data distribution with 4 groups, where 2 groups do not perform better than chance, independent of the threshold chosen, and the other 2 groups perform better for higher values of the threshold. Hence, we achieve Pareto-Fairness for threshold $\geq 9$, whereas, point of equal performance (non-pareto fairness) threshold$=4$, is achieved when all groups perform equal to random chance performance (more details on the distribution is in Section 5.2).
Hence, choosing to minimize Pareto loss is better than minimizing the Parity loss, which can lead to random accuracy.
% Additionally, we restrict to choosing only the Pareto-optimal points which Pareto-dominate the result which minimizes the parity error across all groups. 
% More details on the distribution of this synthetic setup can be found in the supplementary material. 
% \par 
% This approach is similar to current avenues of research which highlight the benefits of fairness through awareness. 
 This approach is similar to current avenues of research where it is acceptable to be aware of the differences \cite{NIPS2018_8035} between various groups' performance in the dataset and operate in a way to improve as opposed to fairness through blindness \cite{pmlr-v80-kilbertus18a}.

\section{Properties of Pareto-Efficient Fairness}

% As noted in Section 3, let the PEF loss over M subgroups be defined as 
% $L_p (o, \hat{o}) = L_{ce}(o, \hat{o}) + \lambda \mathbb{\sum}_{i=0}^M (\alpha \Vert L(i) \Vert_1 + (1-\alpha) \Vert L(i) - \epsilon \Vert_2)$, where o, $\hat{o}$ denotes the true and predicted target variable, $L_{ce}$ denote the cross entropy loss, $L(i)$ denote the pareto loss measured for group $i$, $\lambda$ and $\alpha$ are hyperparameters which control the tradeoff.

In this section, we outline key theoretical results about the Pareto-Efficient Bias Mitigation algorithm's convergence properties, its capacity to discover Pareto curves of subgroup accuracy and their Pareto efficiency. First, we formalize the inherent disalignment between the achieving better accuracy of the target and satisfying the fairness equality constraint for the sensitive group distribution \cite{Menon2018TheCO}. Let $h(X)$ be the class probability function for the target, $t$ be the threshold of class probability for binary classification, $\lambda$ be the parameter that defines the trade-off between accuracy and fairness in the fairness constrained loss function, $B_{\lambda}^*(X)$ denote the Bayes-optimal classifier which minimizes the fairness constrained loss function, then we define $H(\lambda)$ as the disalignment between the target and group distribution.
\begin{align}
    H(\lambda) = E_X[(t - h(X)) . (B_{\lambda}^*(X) - [[h(X) > t)]])
\end{align}

\begin{theorem}
If $H(\lambda) \rightarrow 1$, then under convexity assumptions, minimizing the Group Pareto loss $\mathcal{L}_p$ will converge the classifier to a Pareto Efficient operating point of accuracies $\hat{F} = (\hat{f}_1, \hat{f}_2,..\hat{f}_{|G|})$ , such that for all operating points obtained by strictly enforcing the equality fairness constraint, $F = (f_1, f_2, .. f_{|G|})$, we have $\hat{f}_i \geq f_i$.
\end{theorem}

In the remainder of the section, we present the outline of the proof of the theorem through key lemmas about the \textit{convergence, discoverability and efficiency} of the Pareto-Efficient algorithm from optimization and Pareto optimality theory. 

\subsection{Convergence}
To show that $\mathcal{L}_p$ converges, we use the lemma from \cite{Vincent:2014:SGL:2749482.2749954}, that shows that under block separability of parameters $\beta$, i.e = $\phi(\beta) = \sum_{g=1}^{|G|} \phi^{(g)}(\beta^{(g)})$, where $\beta^{(g)}$ in our case is the Pareto loss for group $g$: $\epsilon_g$, backpropagating using a block level batch gradient descent converges \cite{Tseng2009}.

\begin{lemma}
If f is a convex, twice-differentiable loss function, then the sparse lasso minimizer, min( f + $\lambda \phi (\beta)$), such that $\phi(\beta) = (1-\alpha) \Vert \beta \Vert_2 + \alpha \Vert \beta \Vert_1$, is also convex.
\end{lemma}

In our case, $f$ is the cross-entropy loss function $\mathcal{L}_{ce}$ along with the sparse lasso penalty parameter $\phi (\beta)$ being equal to the Pareto loss penalty term $\alpha \Vert \mathcal{E}_G \Vert_1 + (1 - \alpha) \sigma_G^2 (\mathcal{E}_G)$. \qed

\subsection{Discoverability} 
To show that the converged operating point is Pareto-Efficient, we use the theory of decomposition \cite{MOO} based methods that can discover convex and non-convex Pareto curves by scalarizing $|G|$ multiple objectives - $f_1. f_2.. f_{|G|}$ into a single objective. We minimize the $l_p$ norm of the weighted distance of each objective $f_g$, from a Utopian reference point $f_{opt-g}^*$, i.e.
$min(\big(\mathbb{\sum}_{g=1}^{|G|} w_g |f_g(x) - f_{opt-g}^*|^p \big)^\frac{1}{p})$
% The knowledge of a Utopian reference point $z^*$ is usually based on prior domain knowledge. In our adaptation for Pareto-Efficient Fairness, we have initialized $z^*$ to a vector of subgroup performances when trained exclusively on the subgroup's data alone.

% Using the above Weighted Metric method provides the ability to discover both convex and non-convex pareto curves as shown in \cite{MOO}. Similarly, the $l_\infty$ norm has the ability to discover all points on the Pareto front for some weight vector as stated in the lemma below. \cite{moobook}

% \begin{lemma}
% Let x be a Pareto-optimal solution, then there exists a positive
% weighting w vector such that x is a solution of the weighted
% Tchebycheff problem\\
% $min(\big(\mathop{max}_{m=1}^{M} w_m |f_m(x) - z_m^*| \big))$ ,  where the reference point $z^*$ is the
% utopian objective vector.
% \end{lemma}

However, it can be seen that as $p \rightarrow \infty$, the objective function becomes non-differentiable and hence we want to choose the minimum possible $p$ for which the above statement still holds. While a result for all Pareto curves is still an open problem, we use a significant result from \cite{MOO}, if the Pareto curve is assumed to be continuous.

\begin{lemma}
If the Pareto-front geometry is continuous, where $f_1, f_2, .. f_{|G|}$ can be parameterized as 
$f_1^{p_1} + f_2^{p_2} + .. + f_{|G|}^{p_{|G|}} = C$,
such that $p_i > 0$, for a constant $C$, then for the choice of $p \geq max(p_1, p_2, ... p_{|G|})$, we are guaranteed to discover the Pareto front using the scalarization,
$min(\big(\mathbb{\sum}_{g=1}^{|G|} w_g |f_g(x) - z_g^*|^p \big)^\frac{1}{p})$
\end{lemma}

In finite datasets, we usually make the continuous Pareto curve assumption as it is interpolated using observed points. 
% Under this assumption, the above lemma would hold on the continuous extrapolated Pareto curve. 
Also, in our case, we know that each $f_g \in [0, 1]$, if accuracy (error) is scaled. With this tight bound on the performance values, the condition to be satisfied, $f_1^{p_1} + f_2^{p_2} + .. + f_{|G|}^{p_{|G|}} = C$, becomes trivial to be satisfied empirically for low values of $p_i$ as shown in Figure \ref{pareto-geometry}. 
% For all $p_i = \epsilon$, such that $\epsilon \rightarrow 0^{+}$, we have that $f_i^{p_i} \rightarrow 1$ for $f_i(x) \in [0, 1]$. This is evident as $x^\epsilon \rightarrow 1$, for $x>0$ and $\mathop{lim}_{x \rightarrow 0^+} x^x = 1$. Since all boundary conditions are also bounded by the limit of 1, we can safely assume $C = M$ and satisfy the condition for most practical purposes, as illustrated in Figure \ref{pareto-geometry}. Thus, for choice of $p > \epsilon$, i.e p = {1,2,3..}, we see that our weighted metric method produces all discoverable points on the Pareto curve and hence we can be fairly guaranteed (under the errors of numerical precision) that the minimization procedure would find a point on the Pareto curve. Note that the weights of the weighted metric method in our case is based on the fairness criterion and hence all set to 1. This will further impose the fairness constraints during the discovery of points on the Pareto curve.

% \begin{figure}[htbp]
% 	\begin{center}
% 		\includegraphics[width=3.2in,height=2in]{pareto-curve.pdf} 
%         \setlength{\belowcaptionskip}{-8pt} 
% 		\caption{Pareto geometry condition is satisfied for discoverability of all Pareto-Optimal points for Pareto-Efficient Fairness for low values of p, under the errors tolerated by numerical precision}
% 		\label{pareto-geometry}
% 	\end{center}
% \end{figure} 

\subsection{Efficiency}
We further show using a geometric argument, that if the fairness frontier [Proposition 8 in \cite{Menon2018TheCO}] defined by $H(\lambda)$ for various values of the fairness penalty, $\lambda$, the gain in absolute accuracy obtained by using Pareto-Efficient Fairness is proportional to the value of the absolute gradient of the fairness frontier ($\nabla_\lambda H(\lambda)$), and solely depends on the inherent distribution of the target and groups.
\begin{align}
   | \hat{F} - F | \propto | \nabla_\lambda H(\lambda) |
\end{align}

\begin{figure}[h]
	\begin{center}
		\includegraphics[width=3.2in,height=2.0in]{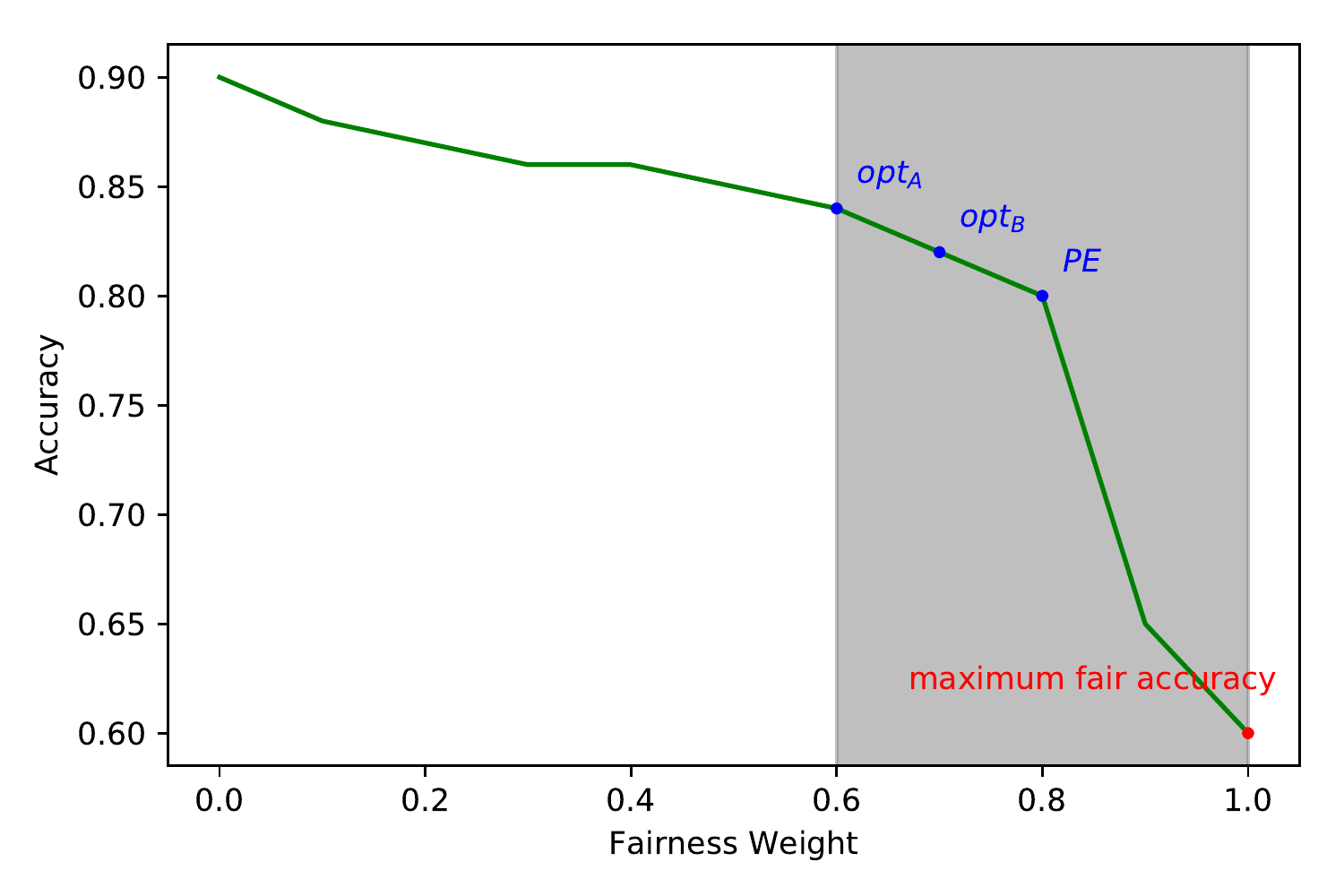} 
        \setlength{\belowcaptionskip}{-8pt} 
		\caption{Relationship between the shape of the fairness frontier and the efficiency gain expected by using PEF. y-axis denotes the maximum achievable overall accuracy for the given fairness constraint (x-axis). Larger values of the fairness constraint do not permit model performances which deviate from the fairness line. If better accuracies are achievable by relaxing the fairness constraint by a small amount, the gain expected by using PEF would also be more.}
		\label{fairness-frontier}
	\end{center}
\end{figure}

% Specifically, if the fairness frontier shows that for small concessions of the fairness requirement ($\Delta \lambda$), the limit of accuracy achievable is much higher ($\Delta_F$), then we have a possibility that PEF would outperform by choosing such a point on the fairness frontier as shown in Figure \ref{fairness-frontier}. However, this is a necessary but not sufficient condition. PEF would choose such a point only if the new point is Pareto-dominating the subgroup accuracies of the operating point without the $\Delta \tau$ fairness concession.
 Specifically, if the fairness frontier shows that for small concessions of the fairness requirement ($\Delta \tau$), the limit of accuracy achievable is much higher ($\Delta u$), then we have a possibility that PEF would outperform by choosing such a point on the fairness frontier as shown in Figure \ref{fairness-frontier}. However, this is a necessary but not sufficient condition. PEF would choose such a point only if the new point is Pareto-dominating the subgroup accuracies of the operating point without the $\Delta \tau$ fairness concession.
The amount of concession in Pareto-dominance that we are willing to allow is domain-dependent and can be controlled by tuning the parameter $\lambda$ in the PEF loss function. Hence, PEF would perform better in conditions where the fairness frontier is steep around the fairness requirement and potential increase in performances are achievable in a Pareto-dominant manner.

% \subsubsection{Multiple metrics of fairness}
% One of the drawbacks of existing subgroup fairness approaches, is that constraints are not applied to multiple metrics of fairness like false positive rate and false negative rates simultaneously. However, the Pareto loss can be scaled easily to mitigate Pareto-loss for several metrics. We achieve this by defining a penalty term per metric of fairness, weighted by separate factors signifying the trade-offs between the multiple Lagrangian relaxation terms. In our evaluation, we accomplish this with an equal weighting between FPR and FNR to overcome the drawbacks of metrics like accuracy which can obfuscate errors in a model.

\section{Related Work}
% Much previous work has explored methodologies for achieving subgroup fairness in ML classification. 
Existing fairness mitigation algorithms often explicitly define constraints on model subgroup performance e.g Equality of Odds \cite{HardtPS16} and enforce using Lagrangian relaxation \cite{Menon2018TheCO,Liu2018,Burke2017MultisidedFF} to achieve corpus level parity across sensitive variables \cite{Zhao2017MenAL}. 
% This work has not been extended to a large number of subgroups. 
With increase in sensitive variables in real case studies, satisfying such strict constraints remain unexplored. \cite{BerkeleyBias,Chiappa2018,Dua:2017}.
%\cite{Zhao2017MenAL} aims to achieve corpus level parity with Lagrangian relaxation. Updates after each batch of training are approximations on samples with the goal of achieving corpus level parity across sensitive variables. The work has not been extended to a large number of subgroups.
%While theoretically this approach can scale to multiple bias features, the empirical behavior for the rate of convergence of the combined loss optimization of the Lagrangian approximation has not been extensively explored.
\cite{Menon2018TheCO} has established that such approximate group fairness constraints are not perfectly possible unless the underlying sub-populations demonstrate perfect accuracy with respect to the target. 
%\cite{Menon2018TheCO} show that a disparate impact constraint is equivalent to a cost sensitive constraint.
%Similarly, they demonstrate for the mean difference (MD) score, the corresponding balanced cost sensitive risk has a cost-parameter that does not depend on $\tau$ (the MD constraint). 
% The work formulates a fairness frontier: for a given lower bound on fairness, they calculate the best excess risk over the solution without a fairness constraint, along with a data dependent theoretical limitation between fairness and accuracy. We extend on these results and show that if the fairness frontier is steep, then PEF achieves better efficiency than existing constrained optimization methods.
\cite{Beutel2017DataDA} models the problem of debiasing as a multi-task learning problem with a penalty if the shared hidden layers of the neural network can be used to predict the sensitive variable accurately. \cite{Zafar2017FromPT} argues about preference based notions of fairness as opposed to ones based on parity. \cite{Samadi:2018:PFP:3327546.3327755} provide fair dimensionality reduction algorithms where bias loss functions are employed. \cite{pmlr-v54-zafar17a} show that on logistic regression and support vector machines, approximate fairness constraints can be enforced with a cost on accuracy.
%In their adversarial approach, the gradients are propagated such that the model is penalized when it predicts the correct sensitive variable and likewise the model is trained to predict the opposite (in case of a binary sensitive variable) of the true label. 
% One potential issue of the model is that it could result in propagating bias in the model in the reverse direction.\par
% Kleinberg ~\cite{Kleinberg2017PlanningWM} considers two biases based on planning (present bias and sunk cost bias) and formulates the reward such that it is parameterized by (b,T) denoting the modification it makes to the final perceived reward at each step. Examples show that behaving naively or in sophisticated manner about one or both of these biases might be optimal in different conditions. The remainder of the paper isn't directly applicable to this work as it is defined as a path finding problem where rewards are adjusted by different biases and defining those parameters can be difficult.

\cite{Pleiss2017} and  \cite{Raghavan2018TheEO} prove that equality of odds cannot be achieved by two models on separate groups which are calibrated 
%(i.e the classification probabilities have a meaning relevant to the population)
, unless both the models achieve perfect accuracy. 
% The work derives a generalized impossibility result that shows that satisfying equalized odds for more than one cost functions which capture FNR/FPR is infeasible.
%Empirically, they show the impact of imposing calibration and an equal cost constraint on real-world datasets. 
The main intuition behind this paper is the hypothesis that similar impossibility regimes exist in real life scenarios, especially when multiple subgroups exist.
% We explore how to avoid unintentional performance degradation in such cases, which only achieve fairness by trivially reducing accuracy to random performance for all subgroups.  

% \subsection{Fairness Metrics}

% In this paper we begin to address the following possible issues with imposing equivalent fairness constraints on sensitive variables subgroups. 

% \begin{enumerate}
	
% 	\item Impossibility Result of optimizing for accuracy across sensitive variable subgroups is incompatible with equalizing False Positive and False Negative rates.
% 	\item Hard constraints even with a relaxation factor can cause inequity across subgroups : equality of odds can't be achieved on subpopulations that vary due to sampling/selection bias, etc.
%     \item Confounding Factors can drastically alter interpretation of fairness across sensitive variables and without their inclusion, we may degrade performance and cause further inequity in supervised learning tasks.
% \end{enumerate}

% \subsection{Pareto-Efficiency}

% \begin{definition}
% \textbf{Pareto Efficiency} is a state where resources are allocated in which it is impossible to redistribute resources to make any one criterion or party better off without making another criterion worse off. 
% \end{definition}

\cite{paretoTradeoff} explores the Pareto optimality between overall accuracy and violation of fairness constraints. 
% Although such a comparison is important and in many cases necessary by a domain expert, it is a measure of two separate metrics and needs to be carefully evaluated. 
However in our work, we focus on the trade-offs between the performance of various comparable subgroups on the Pareto-optimal curve \cite{moobook,MOO}. 
% We extend results from the Pareto-Efficiency literature \cite{moobook,MOO} concerning the discoverability of Pareto curves and show convergence of the Pareto constrained minimization using the properties of subgroup performance metrics. 
Relevant to our work, studies of subgroup specific performance and use of transfer learning like methods have been explored through decoupling in \cite{pmlr-v81-dwork18a}, but do not provide theoretical results on Pareto-Efficiency. To the best of our knowledge, this is the first work which extends strong theoretical results of Pareto-Efficiency to achieve better subgroup performance in data distributions with high disalignment between fairness and accuracy.

%In the same vien, \cite{LiptonEnvyFree}, it was proved that achieving an envy free (fair) allocation of indivisible resources is a co-NP complete problem and as such we heuristic based algorithms need to be employed. Similarly, studies of subgroup specific performance and use of transfer learning like methods have been explored through decoupling in \cite{pmlr-v81-dwork18a}.

% \begin{figure*}[htbp]
% 	\begin{center}
% 		\includegraphics[width=7in,height=2.2in]{impossibility.png} 
%         \setlength{\belowcaptionskip}{-8pt} 
% 		\caption{Example showing impossibility to equalize accuracy, FPR and FNR at the same time}
% 		\label{impossibility}
% 	\end{center}
% \end{figure*}

\section{Evaluation}

We compare our approach with the scaled versions of group fairness \cite{Zhao2017MenAL} and \cite{Beutel2017DataDA} for subgroups. In \cite{Zhao2017MenAL}, the authors optimize for overall accuracy in the constrained setting of ensuring equal false positive rates, but it is generally applicable to other measures of performance. For our comparison we implement an objective to maximize overall accuracy along with a Lagrangian relaxation which adds a penalty for each subgroup that deviates from the overall accuracy. In \cite{Beutel2017DataDA}, the authors implement bias mitigation as a way of erasing the sensitive group membership by back-propagating negative gradients in a multi-headed feedforward neural network. 
% We scale the same for subgroups defined over multiple sensitive variables, where the auxiliary head aims to predict the subgroup class (multi-class classification instead of binary).
We evaluate a comparison of the techniques for both the UCI datasets and synthetic toy data. The UCI Census Adult dataset predicts income category based on demographic information, where the sensitive variables are gender and race in this scenario. 
The UCI German dataset predicts credit type (binary) from demographic information where the sensitive variables are age, gender and personal status. The UCI Heart Attack dataset predicts health status using medical and demographic information, and age and gender are considered sensitive variables. The synthetic dataset is constructed in a manner to illustrate scenarios of how the Pareto loss is useful in skewed datasets.

\subsection{UCI Datasets}
Table \ref{tab:uci_comparison} shows the Pareto-loss, i.e how much each subgroup deviates from the pseudo-optimal of the respective subgroup for the UCI Census Adult dataset. We see that our approach achieves zero Pareto-Loss, while \cite{Zhao2017MenAL} and \cite{Beutel2017DataDA} have non-zero Pareto losses. \cite{Zhao2017MenAL} performs well in terms of lowering the sum of absolute discrepancy of all subgroups' accuracy from the overall accuracy (Parity loss). This is expected as \cite{Zhao2017MenAL} chooses an operating point closest to equal accuracy, when exact equality isn't possible. \cite{Beutel2017DataDA} arrives at an operating point which suffers from non-zero Parity and Pareto-loss. Table \ref{tab:uci_subgroups} clarifies why our approach arrives at a better operating point. We can see that each of the subgroups have better individual accuracy than all the other approaches, some even better than the baseline. This confirms empirically that our objective function matches (and sometimes exceeds due to transfer learning) the heuristic pseudo-optimal performance for each subgroup (last row of Table \ref{tab:uci_subgroups}). Similar performance improvement was also observed on the UCI German and Heart Attack datasets and the condensed results are shown in Tables \ref{tab:german_comparison} and \ref{tab:heart_comparison} respectively.

% \resizebox{\columnwidth}{!}{%
\begin{table}[tp]
\parbox{\linewidth}{
%\footnotesize

\caption{UCI Adult dataset with bias mitigation algorithms}
\begin{center}
\resizebox{0.60\textwidth}{!}{%
\begin{tabular}{lccccc}
\hline
\textbf{Model} & \textbf{Accuracy} & \textbf{FPR} & \textbf{FNR} & \textbf{Parity Loss} & \textbf{Pareto Loss}\\\hline
Baseline (no bias loss) & 0.630 & 0.253 & 0.747 & 0.199 & 0.016\\\hline
\cite{Zhao2017MenAL} & 0.619 & 0.283 & \textbf{0.712} & \textbf{0.167} & 0.133\\
\cite{Beutel2017DataDA} & 0.648 & 0.224 & 0.769 & 0.226 & 0.077\\
Pareto-Efficient Loss & \textbf{0.678} & \textbf{0.165} & 0.830 & 0.250 & \textbf{0.000}\\\hline
\end{tabular}}
\end{center}
\label{tab:uci_comparison}
%\normalsize
}
\hfill
\parbox{\linewidth}{
%\footnotesize
\caption{Subgroup Accuracy on UCI Adult dataset}
\begin{center}
\resizebox{0.48\textwidth}{!}{%
\begin{tabular}{lccccc}
\hline
\textbf{Model} & \textbf{Subgroup 1} & \textbf{2} & \textbf{3} & \textbf{4} & \textbf{Pareto Loss}\\\hline
Baseline (no bias loss) & 0.890 & 0.883 & 0.818 & 0.784 & 0.016\\\hline
\cite{Zhao2017MenAL} & 0.853 & 0.856 & 0.806 & 0.778 & 0.133\\
\cite{Beutel2017DataDA} & 0.882 & 0.872 & 0.824 & 0.780 & 0.077\\
Pareto-Efficient Loss & \textbf{0.935} & \textbf{0.915} & \textbf{0.844} & \textbf{0.797} & \textbf{0.000}\\
Subgroup Pareto Frontier & 0.934 & 0.894 & 0.815 & 0.783 & N/A \\\hline
\end{tabular}}
\end{center}
\label{tab:uci_subgroups}
%\normalsize
}
\end{table}%

\begin{table}[tp]
\parbox{0.49\linewidth}{
\caption{UCI German dataset with bias mitigation algorithms}
\begin{center}
\resizebox{3cm}{!}{%
\begin{tabular}{lc}
\hline
\textbf{Model} & \textbf{Accuracy}\\\hline
Baseline (no bias loss) & 0.749\\\hline
\cite{Zhao2017MenAL} & 0.696\\
\cite{Beutel2017DataDA} & 0.694\\
Pareto-Efficient Loss & \textbf{0.711}\\\hline
\end{tabular}}
\end{center}
\label{tab:german_comparison}
}
\hfill
\parbox{0.49\linewidth}{
\caption{UCI Heart Attack dataset with bias mitigation algorithms}
\begin{center}
\resizebox{3cm}{!}{%
\begin{tabular}{lc}
\hline
\textbf{Model} & \textbf{Accuracy}\\\hline
Baseline (no bias loss) & 0.939 \\\hline
\cite{Zhao2017MenAL} & 0.870\\
\cite{Beutel2017DataDA} & 0.837\\
Pareto-Efficient Loss & \textbf{0.879}\\\hline
\end{tabular}}
\end{center}
\label{tab:heart_comparison}
}
\end{table}%

\subsection{Synthetic Data}
We varied the hyper-parameters that define the synthetic subgroup distributions and found that for cases where it is possible for all subgroups to achieve same level of accuracy, we see that our approach remains similar in subgroup performance to \cite{Zhao2017MenAL} and \cite{Beutel2017DataDA}. However, for the use cases where subgroups differ in their ``pseudo-optimal'' performance, \cite{Zhao2017MenAL} fails to achieve the ``Pareto'' accuracy for all subgroups and hence results in lower overall accuracy. \cite{Beutel2017DataDA} however continues to perform well, and our approach only shows improvements in a few scenarios. We provide detailed performance numbers and subgroup distribution parameters on the synthetic cases below.

We use a joint data distribution for binary sensitive variables $a$,$b$ and $d$, a confounding variable $C$ introduced to control the alignment between the target label $T_0$ and sensitive group distribution. We present the various dependence between the variables $C$ and the sensitive variables and the corresponding performance in Table \ref{tab:synthetic_subgroups} for minimizing the Pareto loss compared to minimizing Parity Loss \cite{Zhao2017MenAL} and adversarial losses \cite{Beutel2017DataDA}. \\

\begin{table}[t!]
\footnotesize
\caption{Change in overall accuracy as compared to baseline (no bias loss) on synthetic dataset}
\begin{center}
\resizebox{0.60\textwidth}{!}{%
\begin{tabular}{lccc}
\hline
\textbf{\vtop{\hbox{Confounding dependency}\hbox{$\mu_C: T_0 = Bern(\mathcal{N}(\mu_C, \sigma^2)) $}}} & \textbf{Pareto} & \textbf{\cite{Zhao2017MenAL}} & \textbf{\cite{Beutel2017DataDA}}\\\hline
2*a + 1*b & 0 & 0 & 0\\
2*b - 2*a & 0 & 0.02 & 0.02\\
4*b & 0 & 0 & -0.002\\
8*b & 0 & -0.06 & 0\\
2*a + 2*b + 2*d & 0 & 0 & 0\\
\vtop{\hbox{\strut (a,b): \{(0,0): 3, (0,1): 11,}\hbox{\strut (1,0): 4, (1,1): 8\}}}
 & 0 & 0 & 0\\
\vtop{\hbox{\strut (a,b): \{(0,0): 3, (0,1): 1,}\hbox{\strut (1,0): 4, (1,1): 8\}}}
 & 0 & -0.04 & -0.04\\
\vtop{\hbox{\strut (a,b): \{(0,0): 3, (0,1): 11,}\hbox{\strut (1,0): 4, (1,1): 9\}}}
 & 0 & -0.01 & -0.01\\\hline
\end{tabular}}
\end{center}
\label{tab:synthetic_subgroups}
\normalsize
\end{table}%

\subsubsection{Skewed Dataset Scenario}
Below, we show the distribution that was used to generate a skewed dataset where it is possible that attempting to achieve strict equality might result in trivial accuracy. We use a joint data distribution for binary sensitive variables $A$ and $B$, a threshold variable $C$ introduced to classify the target label $T_0$. Let A, B be Bernoulli random variables,

\begin{figure}[h]
	\begin{center}
		\includegraphics[width=3.2in,height=2in]{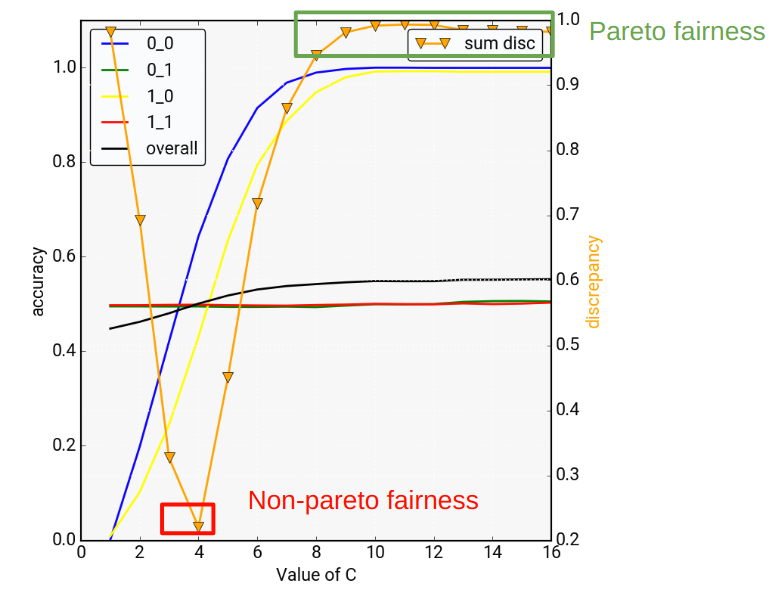}
        \setlength{\belowcaptionskip}{-8pt} 
		\caption{Synthetic dataset where Pareto Efficient fairness is desirable}
		\label{pareto}
	\end{center}
\end{figure}

\begin{center}
\begin{align}
    A \sim Bern(p), B \sim Bern(p), p=.5\\
    Bern(p) = f(k;p) =
    \left\{
        \begin{array}{cc}
                p & \mathrm{if\ } k=1 \\
                1-p & \mathrm{if\ } k=0 \\
        \end{array} 
    \right.
\end{align}
with a selected constant $0<c<p$
\[ C = 
\begin{cases}
\mathcal{U}(0,1) & if B = 1\\
\mathcal{N}(p, \sigma^2) - c & if B = 0, A= 0\\
\mathcal{N}(p, \sigma^2) + c & if B = 0, A= 1 \\
\end{cases}
\]

$T_0 \sim Bern(C)$ \\
\end{center}

In this scenario, there are two subgroups for which $B=0$, the confounding variable $C$ has uniform random values and for the two subgroups which have $B=1$, the confounding variable $C$ is distributed normally with means at two different values separated by a constant. Suppose we were to train a bias mitigation model $M$, which uses only the perceived non-sensitive variable "C" as the feature to classify to identify target label $T_0$  in order to achieve equal performance in terms of accuracy. While trying to increase overall accuracy, there is a potential trade-off where the algorithm could penalize two subgroups identified by $B=0$ (0\_0, 1\_0), without any gain in the subgroups identified by $B=1$ (0\_1, 1\_1). Effectively, this implies that two subgroups are penalized to ensure that accuracy is close to the uniformly random accuracy as in the other two subgroups as shown in Figure \ref{pareto}. This is counter-productive for each of the subgroup's performance in question. Hence, it is necessary to guard against such scenarios, by explicitly accounting for this edge-case.

\subsection{Bias Loss Weights:}
\begin{figure}[h]
\centering
	\includegraphics[width=0.45\linewidth]{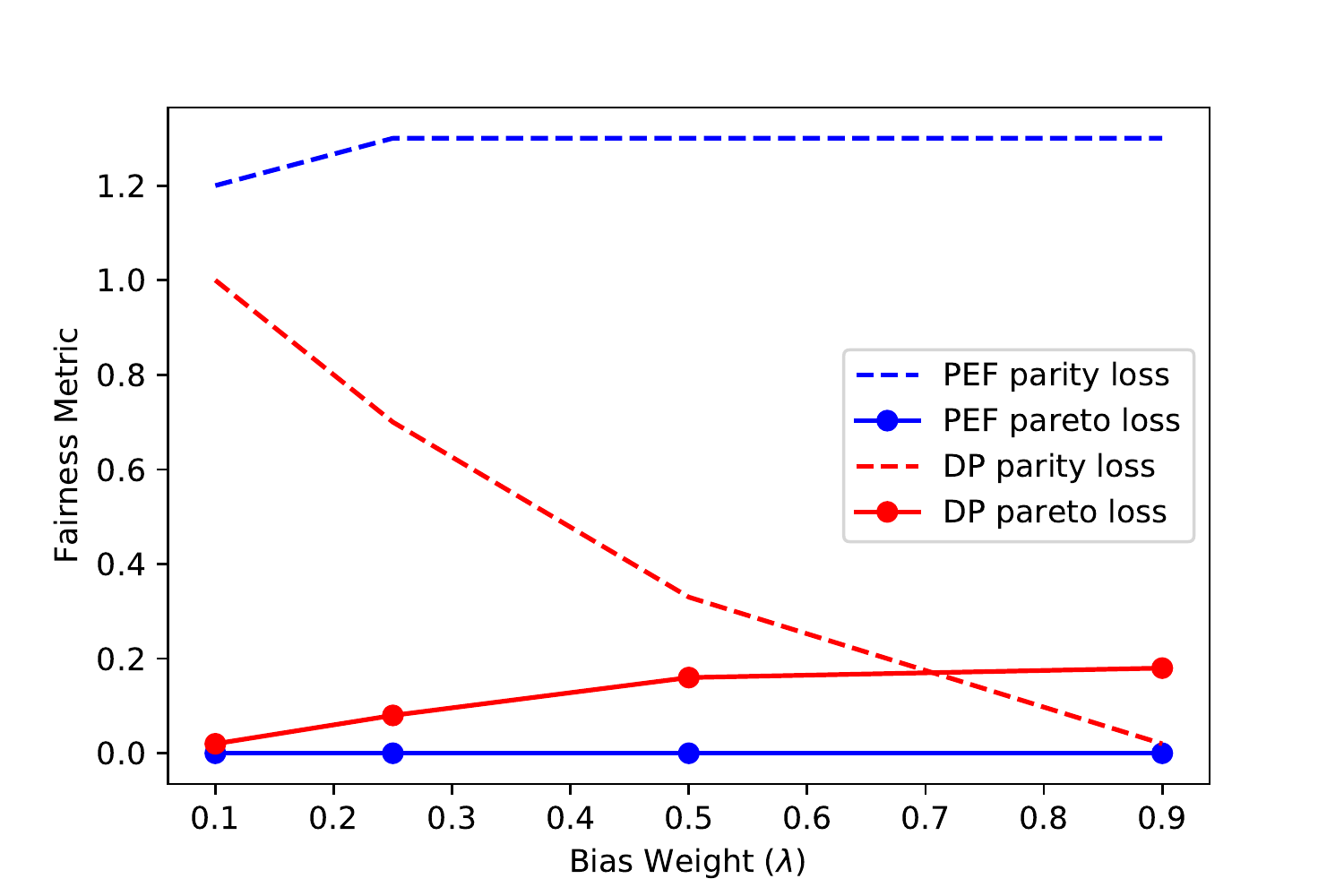} 
    \setlength{\belowcaptionskip}{-8pt} 
	\caption{Comparison of effects of bias-weights on Pareto loss and Parity loss between strict enforcement (Demographic Parity- DP) and Pareto-Efficient Fairness (PEF) methodologies. In case of DP, as the weight ($\lambda$) increases, parity loss decreases but at the cost of significant Pareto loss increase. However for PEF, increase in weight, reduces Pareto loss without significantly impacting the parity loss.}
	\label{fig:1a}
\end{figure}

The Pareto-Loss function contains a single bias loss weight term ($\lambda$, although it can be a vector as per domain) which controls the trade off between achieving overall accuracy and Pareto-Efficient levels for each of the subgroups with equal penalty per subgroup. Modifying this weight demonstrates the difference between the equalized odds loss and the Pareto loss as can be seen in Figure \ref{fig:1a}. As the bias weight is increased, the Pareto loss bias mitigation moves the threshold towards Pareto-Efficient levels with higher overall accuracy, whereas equalized loss moves towards the operating point where discrepancy across subgroups is minimized, even thought it doesn't improve accuracy of at least one of the subgroups.
%It is possible to weight the discrepancies of each of the subgroups differently or apply a constant as warranted by domain expertise of the practitioner. 

\begin{figure}[h]
\centering
	\includegraphics[width=0.45\linewidth]{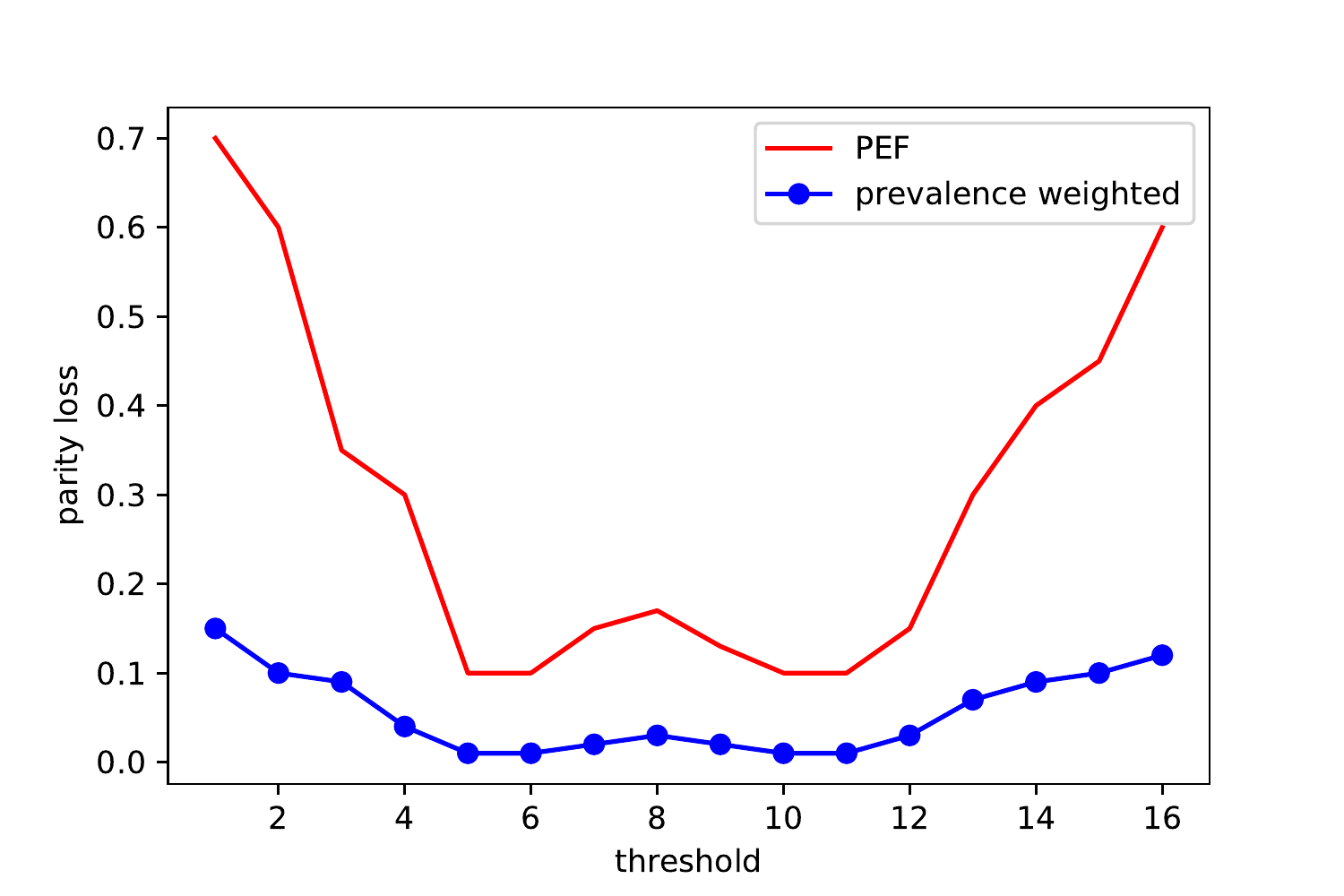} 
    \setlength{\belowcaptionskip}{-8pt} 
	\caption{Weighing fairness loss by prevalence (blue) provides a deceptive notion of fairness by ignoring the minority subgroup and lowering the overall parity and Pareto loss, without actually providing any benefit to the minority subgroup. Hence, we adopt an equal weighting of subgroup level Pareto losses in PEF.}
	\label{fig:1b}
\end{figure}

\subsection{Prevalence}
The number of samples in each subgroup varies and as such some minority population subgroups could be ignored during some bias mitigation techniques. The approach mentioned in \cite{Kearns2017PreventingFG} defines discrepancies weighted by the population ratio of each of the subgroups. This may have perverse effects on minority subgroups whose performance deviates from either the overall performance or the Pareto-Efficient level. The contribution of the minority subgroup may be ignored while optimizing their loss function. This is clearly seen in Figure \ref{fig:1b} where the curve representing the discrepancy from the overall performance can be seen to have lowered and smoothened when multiplied with the subgroup's population ratio. In some domains, this might not be the intended fairness criterion and can be quite deceiving when the majority subgroup, which also guides the overall performance dominates the sum of discrepancy loss too. As such, the authors guard against the tyranny of the majority, by weighting subgroups equally, independent of population prevalence. This can be modified to appropriate weights by the domain practitioner.

\subsection{Model Capacity}
We used 3 layers of feed forward networks with 256, 128 and 64 neurons fully connected in all our comparisons of relevant losses. However, we did notice that the difference in the losses varied with the size of the model used as noted in Table \ref{tab:uci_model_size}. The gains achieved from Pareto-Loss is higher when the model capacity increases, as it is able to capture the subgroup specific performance updates with a larger number of parameters.

\begin{table}[h]
\footnotesize
\caption{Effect of model size on subgroup accuracy (compared with baseline) on UCI Adult dataset}
\begin{center}
\resizebox{0.60\textwidth}{!}{%
\begin{tabular}{lcccc}
\hline
\textbf{Model size} & \textbf{Subgroup 1} & \textbf{2} & \textbf{3} & \textbf{4}\\\hline
64 & 0.923 (0.929) & 0.898 (0.914) & 0.812 (0.849) & 0.733 (0.809)\\
$128,64$ & 0.902 (0.925) & 0.884 (0.909) & 0.820 (0.849) & 0.781 (0.805)\\
$256,128,64$ & \textbf{0.935} (0.890) & \textbf{0.915} (0.883) & \textbf{0.844} (0.818) & \textbf{0.797} (0.784)\\\hline
\end{tabular}}
\end{center}
\label{tab:uci_model_size}
\normalsize
\end{table}%

\section{Conclusion}
Real-world datasets often display subgroup population skew. To mitigate in the construction of fairness-aware classifier, we utilize softer Pareto-efficiency fairness constraints. When subgroup populations contain deviations in prevalence and underlying distributions, a Pareto-Efficient approach yields better overall and individual subgroup performance when compared to other bias-mitigation algorithms enforcing hard equality constraints. The approach is appropriate for classification problems containing multiple sensitive variables. As demonstrated, the proposed methodology does not degrade performance in terms of accuracy for datasets who do demonstrate balanced distributions across subgroups.  In fact, we demonstrated a substantial increase in global accuracy and individual subgroup accuracy on three UCI datasets as compared to existing fairness algorithms with hard equality constraints. \par
    
We note that our approach requires computing subgroup pseudo-optimal accuracy on sufficient subgroup samples. 
% This requirement will grow with the number of assigned sensitive variables. 
Avoiding the pitfalls of achieving trivial accuracy with more sensitive subgroups comes at this cost of pre-processing/evaluation time.
% Similarly, the paper's approach is predicated on examining the individual subgroup populations. 
As the number of sensitive variables grow, the sample size required to establish pseudo-optimal statistics may be infeasible and individual fairness definitions \cite{pac-fair, DBLP:journals/corr/abs-1807-00787} may be appropriate.
% For example, it is within reason, to consider a scenario where a subgroup can be reduced to a possible sample size of 1. 
% This scenario is not addressed in this paper, but is of considerable importance when exploring sensitive variable constraints in ML as applied in policy decisions.  

%\pagebreak

\bibliographystyle{plain}
\bibliography{sample-bibliography}

\section{Appendix A}

\subsection{Convergence}
We first present the lemma derived for sparse lasso regularizers \cite{Vincent:2014:SGL:2749482.2749954}, that show the convexity of block regularized minimizers, which also minimize the loss function under model parameter ($\beta$) constraints.
\begin{lemma}
If f is a convex, twice-differentiable loss function, then the sparse lasso minimizer, min( f + $\lambda \phi$), such that $\phi(\beta) = (1-\alpha) \Vert \beta \Vert_2 + \alpha | \beta |$, is also convex.
\end{lemma}

Moreover, it has be shown that the convexity argument holds even when $\beta \in \mathbb{R}^n$, as long as the block separability of $\beta$ holds, i.e = $\phi(\beta) = \sum_{i=0}^M \phi^{(i)}(\beta^{(i)})$, where $i^{th}$ component of $\beta$ denotes the $i^{th}$ subgroup's performance's deviation from their group optimal performance. Hence, adopting a block level gradient descent, where the gradients are backpropagated only after each batch's block performances are computed has been shown to converge in \cite{Tseng2009}. 

\subsection{Discoverability}

The above section shows that the Pareto-Efficient Fairness loss indeed converges to a minima with the use of a convex loss function. However, it remains to be seen that the minima obtained is a pareto-optimal operating point. For this, we now provide insights behind the choice of the regularizers in Pareto-Efficient Fairness, based on the theory of multiple objective optimization \cite{MOO}. Specifically, we use the theory of decomposition based methods which employ a scalarization technique to convert $M$ multiple objectives - $f_1. f_2.. f_M$ into a single objective using a Weighted Metric method. Here, the distance of each objective from a Utopian reference point is measured and a corresponding $l_p$ norm is minimized, i.e.
$min(\big(\mathbb{\sum}_{m=1}^{M} w_m |f_m(x) - z_m^*|^p \big)^\frac{1}{p})$
\\
The knowledge of a Utopian reference point $z^*$ is usually based on prior domain knowledge. In our adaptation for Pareto-Efficient Fairness, we have initialized $z^*$ to a vector of subgroup performances when trained exclusively on the subgroup's data alone.

Using the above Weighted Metric method provides the ability to discover both convex and non-convex pareto curves as shown in \cite{MOO}. Similarly, the $l_\infty$ norm has the ability to discover all points on the Pareto front for some weight vector as stated in the lemma below. \cite{moobook}

\begin{lemma}
Let x be a Pareto-optimal solution, then there exists a positive
weighting w vector such that x is a solution of the weighted
Tchebycheff problem\\
$min(\big(\mathop{max}_{m=1}^{M} w_m |f_m(x) - z_m^*| \big))$ ,  where the reference point $z^*$ is the
utopian objective vector.
\end{lemma}

However, it can be seen that as $p \rightarrow \infty$, the objective function becomes non-differentiable and hence it is of interest to us that we choose the minimum possible p for which the above statement still holds. While a universal result for all Pareto curves is still unknown, a significant result from \cite{MOO} is presented below, if the Pareto curve is known to be continuous.

\begin{lemma}
If the Pareto-front geometry is continuous, where $f_1, f_2, .. f_M$ denote the $M$ objectives to be optimized which can be parameterized as 
$f_1^{p_1} + f_2^{p_2} + .. + f_M^{p_M} = C$,
such that $p_i > 0$, for a constant $C$, then for the choice of $p \geq max(p_1, p_2, ... p_M)$, the same guarantees of discoverability from the Tchebycheff problem will hold when using the scalarization,
$min(\big(\mathbb{\sum}_{m=1}^{M} w_m |f_m(x) - z_m^*|^p \big)^\frac{1}{p})$
\end{lemma}

In real datasets, the number of points observed on the Pareto front is finite, and hence we usually make the assumption that the Pareto curve is extrapolated using the observed points. Under this assumption, the above lemma would hold on the continuous extrapolated Pareto curve. Also, in our case where we optimize subgroup performance, we know that each $f_i \in [0, 1]$, if accuracy (error) or any other performance metric is scaled. With this tight bound on the performance values, the condition to be satisfied, $f_1^{p_1} + f_2^{p_2} + .. + f_M^{p_M} = C$, becomes trivial to be satisfied empirically under the constraints of numerical precision. For all $p_i = \epsilon$, such that $\epsilon \rightarrow 0^{+}$, we have that $f_i^{p_i} \rightarrow 1$ for $f_i(x) \in [0, 1]$. This is evident as $x^\epsilon \rightarrow 1$, for $x>0$ and $\mathop{lim}_{x \rightarrow 0^+} x^x = 1$. Since all boundary conditions are also bounded by the limit of 1, we can safely assume $C = M$ and satisfy the condition for most practical purposes, as illustrated in Figure \ref{pareto-geometry}. Thus, for choice of $p > \epsilon$, i.e p = {1,2,3..}, we see that our weighted metric method produces all discoverable points on the Pareto curve and hence we can be fairly guaranteed (under the errors of numerical precision) that the minimization procedure would find a point on the Pareto curve. Note that the weights of the weighted metric method in our case is based on the fairness criterion and hence all set to 1. This will further impose the fairness constraints during the discovery of points on the Pareto curve.

\subsection{Efficiency}
In this subsection, we provide an analysis of when we expect PEF to outperform standard notions of fairness, like equality of opportunity, i.e when PEF has higher efficiency. \cite{Menon2018TheCO} defines the fairness-frontier which intuitively measures the trade-off  between utility ($u$) (accuracy) and fairness ($\tau$)
in the distribution inherent to the problem, rather than one owing to the
specific technique one uses, no matter how sophisticated
it may be, by computing the fundamental limits of what accuracy
is achievable by any classifier. Specifically, the frontier is computed using cost-sensitive measure which quantifies the alignment between the Bayes-optimal plug-in classifier thresholds for the outcome and sensitive attribute distributions. [Proposition 8 in \cite{Menon2018TheCO}].
\begin{lemma}
As the absolute gradient of the fairness frontier increases near the desired fairness constraint, the efficiency gained from using PEF is monotonically non-decreasing. 
\end{lemma}

Specifically, if the fairness frontier shows that for small concessions of the fairness requirement ($\Delta \tau$), the limit of accuracy achievable is much higher ($\Delta u$), then we have a possibility that PEF would outperform by choosing such a point on the fairness frontier as shown in Figure \ref{fairness-frontier}. However, this is a necessary but not sufficient condition. PEF would choose such a point only if the new point is Pareto-dominating the subgroup accuracies of the operating point without the $\Delta \tau$ fairness concession.
The amount of concession in Pareto-dominance that we are willing to allow is domain-dependent and can be controlled by tuning the parameter $\lambda$ in the PEF loss function. Hence, PEF would perform better in conditions where the fairness frontier is steep around the fairness requirement and potential increase in accuracies are achievable in a Pareto-dominant manner.

\end{document}